\documentclass[final]{cvpr}

\usepackage{mycvpr}

\def\cvprPaperID{5873} 
\def\confYear{CVPR 2021}

\begin{document}

\title{Deep Implicit Moving Least-Squares Functions for 3D Reconstruction}
\author{Shi-Lin Liu$^{1,3}$ \hspace{2mm} Hao-Xiang Guo$^{2,3}$ \hspace{2mm} Hao Pan$^{3}$ \hspace{3mm} Peng-Shuai Wang$^{3}$ \hspace{2mm} Xin Tong$^{3}$ \hspace{2mm} Yang Liu$^{3}$\\[1mm] \small
$^1$University of Science and Technology of China \hspace{2mm}
$^2$Tsinghua University\hspace{2mm}
$^3$Microsoft Research Asia\\
\tt \small freelin@mail.ustc.edu.cn ghx17@mails.tsinghua.edu.cn \\ \tt \small \{haopan,penwan,xtong,yangliu\}@microsoft.com
}


\maketitle

\begin{abstract}

  Point set is a flexible and lightweight representation widely used for 3D deep learning. However, their discrete nature prevents them from representing continuous and fine geometry, posing a major issue for learning-based shape generation. In this work, we turn the discrete point sets into smooth surfaces by introducing the well-known implicit moving least-squares (IMLS) surface formulation, which naturally defines locally implicit functions on point sets.  We incorporate IMLS surface generation into deep neural networks for inheriting both the flexibility of point sets and the high quality of implicit surfaces. Our IMLSNet predicts an octree structure as a scaffold for generating MLS points where needed and characterizes shape geometry with learned local priors.  Furthermore, our implicit function evaluation is independent of the neural network once the MLS points are predicted, thus enabling fast runtime evaluation. Our experiments on 3D object reconstruction demonstrate that IMLSNets outperform state-of-the-art learning-based methods in terms of reconstruction quality and computational efficiency. Extensive ablation tests also validate our network design and loss functions.

\end{abstract}

\section{Introduction} \label{sec:intro}

Point set is probably the most widely used representation for 3D deep learning.
Compared with other 3D representations like polygonal meshes and volumetric grids,
point sets are naturally embedded as neurons in DNNs, easy to acquire, have minimal extra structure to maintain, capture complex geometry and topology dynamically, and induce no wasted computation for free-space regions.
Indeed, point sets have been used in the deep learning-based 3D analysis for diverse tasks \cite{Qi2016,qi2017pointnetplusplus,Guo2019Survey}.
However, when using point sets for the generation of 3D data by deep learning, we have the flexibility to model changing topology and complex surfaces on one hand, but also suffer from the discrete and rough geometry on the other hand.

Recent works have focused on generating shapes in the forms of meshes and polygonal patches, but their shape representation abilities are still restricted by their discrete and non-smooth nature. The deep implicit function approaches \cite{Mescheder2019,Chen2019,Park2019}, instead, define smooth functions on entire 3D domains to guarantee result continuity, and have shown to be promising for high-quality 3D reconstruction. However, the implicit surface generation is inefficient, because for each point in the 3D domain the network has to be evaluated individually before the surface can be extracted.
In this paper, we combine the advantages of both the implicit function approaches and the point set methods, by extending the point set representation to model implicit surfaces for high-quality 3D generation, while preserving the inherent flexibility and computational efficiency of explicit point sets.

For modeling smooth surfaces via point sets, we adopt point set surfaces~\cite{Alexa2001} and use moving least-squares (MLS) interpolation of the points to define locally implicit functions over a narrow band region of the point set.
Specifically for the implicit MLS formulation \cite{Kolluri2008} used in this paper, for any spatial point inside the narrow region, the implicit MLS function maps it to a signed distance value to the zero level set surface, defined by the weighted blending of signed distances to oriented planes supported at the nearby points; the zero level set surface can then be extracted as a smooth and continuous surface for shape representation.

\begin{figure}[t]
    \centering
    \includegraphics[width=1\linewidth]{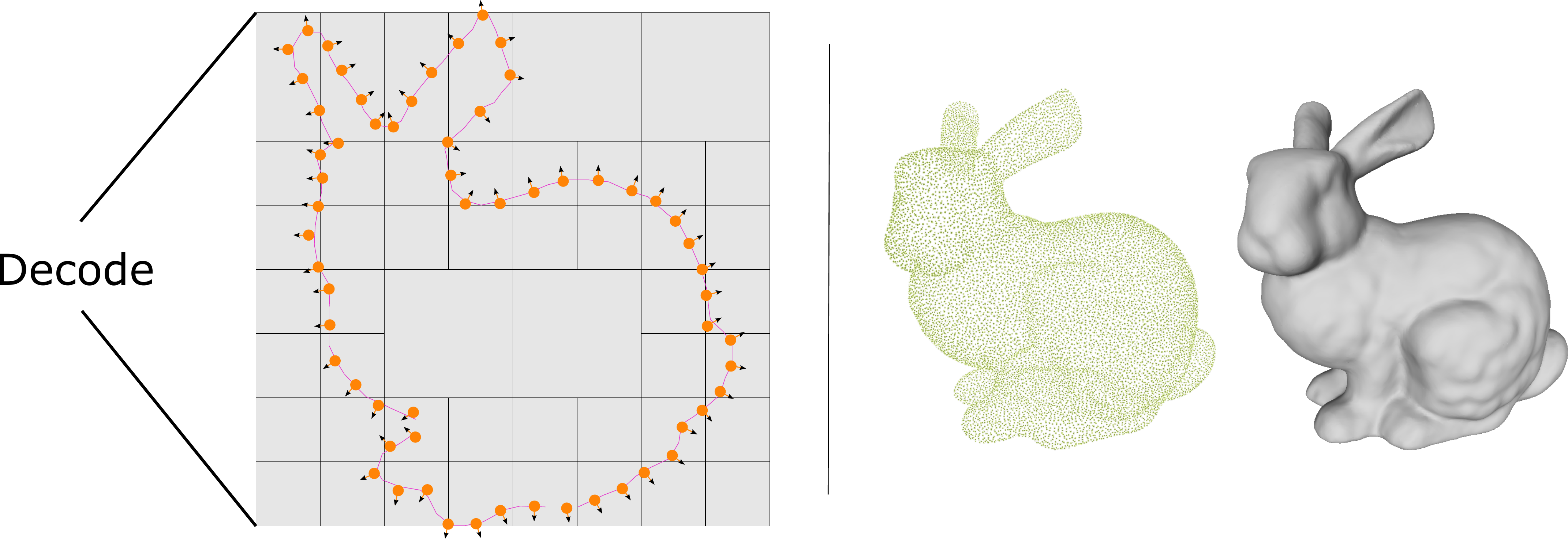}
    \caption{Illustration of our deep implicit MLS surface generation method. Left: the network decoder produces a set of oriented points scaffolded by an octree (2 points for each non-empty leaf octant in this example), which defines an implicit zero level set surface that is the target shape for shape generation. Right: an example of the reconstructed bunny shape from 3D points via an implicit MLS.}
    \label{fig:2d_scheme}
    \vspace{-4mm}
\end{figure}

While MLS surfaces have been well studied in 3D reconstruction and rendering~\cite{Cheng2008},
to incorporate the representation into a deep learning framework poses novel challenges and opportunities unseen in existing point-set-based or implicit-representation-based methods.
First, the points define implicit MLS surfaces most effectively when they are dense enough and distributed evenly over the shapes to reconstruct.
While most point generation approaches fix the number of points and consume many resources for the straightforward prediction of dense points that are hard to generalize, we introduce an \emph{octree-based scaffolding} for generating variable numbers of MLS points only where needed according to the target shape (\cf \cref{fig:2d_scheme}), and further regularize the point distribution via tailored loss functions.
Second, to measure the predicted implicit functions for both training supervision and test evaluation, while existing implicit methods \cite{Mescheder2019,Chen2019,Park2019,Peng2020} have to use dense sampling over the whole 3D domain, the MLS surface is naturally localized inside the narrow band region of generated points, which prompts us to use more succinct sampling only at the octree nodes for supervision and evaluation.
Also, the evaluation is independent of the network once all MLS points are embedded in the 3D domain, thus avoiding the costly per-point network evaluation that is typical of other implicit methods.

We use extensive ablation tests to validate the design choices made.
We also demonstrate that our deep implicit MLS surface approach outperforms both the point set generation methods and other global or local implicit function methods, through the 3D object reconstruction task.

\section{Related Work} \label{sec:related}

\myparagraph{Deep representations for 3D generation}
Point set is a popular representation utilized by many works~\cite{Su2017,Achlioptas2018,Yifan2019,PointFlow}. However, as the number of points is usually specified, its power for representing detailed geometry is restrained and needs further point upsampling~\cite{Yu2018,Wang2019} to improve the point density and shape quality.

Dense voxels \cite{Choy2016,Brock2016,Wu2016} represent shape occupancy well, but its high memory cost precludes its usage for representing 3D contents with high-resolution. Sparse voxels including octrees~\cite{Riegler2017,Wang2017,Tatarchenko2017,Hane2017,choy20194d} overcome these issues with great efficiency in both memory and computational aspects.

Mesh representations~\cite{Wang2018,kato2018renderer} and patch-based representations~\cite{Groueix2018,Deprelle19,Wang2018a,Yang2018,deepgeomprior} are convenient 3D representations for improving shape quality. However, their abilities are constrained by either the predefined mesh topology and resolution, or the disconnectivity of multiple patches. Intermediate 3D representations like coarse voxels~\cite{Gkioxari2019} or shape skeleton~\cite{Tang2019} are possible ways to further enhance their quality. 

Primitive-based representations use a set of simple geometric objects like planes~\cite{Liu2018a} and cuboids~\cite{Tulsiani2017,PRNN2017,Sun2019} to approximate 3D shapes.  Structure-based representations~\cite{Li2017,Niu2018} explicitly encode semantic part structures as cuboids and reconstruct parts within each cuboid using voxel representations.  Although they are suitable to characterize shape structures, their approximation quality is also limited due to the simplicity of primitives.

Recently implicit surface-based deep learning approaches \cite{Chen2019,Mescheder2019,Park2019,Atzmon2020} offer a smooth and continuous 3D representation and enable functional evaluation in a continuous space. For a given point, the network predicts its occupancy or the signed distance from it to the surface.  These techniques are further improved recently by incorporating local features to model more detailed shape geometry  \cite{Peng2020,Chabra2020,Chibane2020,Jiang2020}.

Our approach belongs to the deep implicit category by modeling smooth and continuous implicit MLS surfaces, while also enjoying the flexibility and efficiency of explicit point set generation; it is a hybrid 3D deep learning representation that combines advantages from both point sets and implicit functions.

\myparagraph{Surface reconstruction from point clouds}
Surface reconstruction techniques have been studied for several decades (\cf the comprehensive survey~\cite{Berger2017}). Among them, a set of methods impose global or local smoothness priors for reconstructing high-quality results from point clouds, including multi-level partition of unity (MPU)~\cite{Ohtake2003}, Poisson reconstruction~\cite{Kazhdan2006,Kazhdan2013}, radial basis functions (RBF)~\cite{Carr2001}, and moving least-squares surfaces (MLS)~\cite{Levin1998,Alexa2001,Guennebaud2007} which are widely used for point set surface modeling and rendering \cite{Pauly2002}. Due to the fast and local evaluation property of MLS, we choose MLS surfaces as our deep 3D representation.  MLS surfaces can be classified into two types~\cite{Cheng2008}: projection MLS surfaces and implicit MLS surfaces (IMLS). The former is defined by a set of stationary points via iterative projection, while the latter defines an implicit function directly. We use IMLS for incorporating signed distance supervision easily and enabling fast function evaluation.

\section{Method} \label{sec:method}

We first review the implicit MLS surface (IMLS) definition and present our deep IMLS design in \cref{subsec:imls,subsec:dimls}, then we detail the network design and loss functions in \cref{subsec:net,subsec:loss}.

\subsection{IMLS surface} \label{subsec:imls}
The implicit MLS surface~\cite{Kolluri2008} is defined as follows.
Denote $\mP = \{\mp_i \in \mathbb{R}^3\}_{i=1}^N$ as a set of 3D points and each point is equipped with unit normal vector $\mn_i \in \mathbb{R}^3 $ and control radius $r_i \in \mathbb{R}^+$. For convenience, we call these points as \emph{MLS points}.

For each MLS point $\mp_i$, a signed distance function from $\mx \in \mathbb{R}^3$ to its tangent plane is defined as $\left<\mx - \mp_i, \mn_i\right>$,  where $\left<\cdot,\cdot\right>$ is the inner product.  By weighted averaging of all point-wise signed distance functions, we have an implicit function $F(\mx)$ whose zero level set defines the implicit surface $\mS$:
\begin{equation}\label{eq:mls}
    F(\mx) :=  \dfrac{\sum_{\mp_i \in \mP} \theta(\|\mx - \mp_i\|, r_i) \cdot \left< \mx - \mp_i, \mn_i \right>}{\sum_{\mp_i \in \mP} \theta(\|\mx - \mp_i\|, r_i)}.
\end{equation}
Here we set the weight function as $\theta(d, r) = \exp(-d^2/r^2)$.
Kolluri proved that \emph{under a uniform sampling condition, the IMLS surface $\mS$ is a geometrically and topologically correct approximation of the original surface where $\mP$ are sampled from, and the IMLS function $F$ is a tight approximation of the signed distance function of the original surface} \cite{Kolluri2008}.

As the weight function decays when $\mx$ is  away from $\mp_i$, the evaluation of $F(\mx)$ can be accelerated by considering nearby MLS points only. \cref{eq:mls} can be revised as:
\begin{equation}\label{eq:mls2}
    F(\mx) :=  \dfrac{\sum_{\mp_i \in \Omega(\mx)} \theta(\|\mx - \mp_i\|, r_i) \cdot \left<\mx - \mp_i, \mn_i \right>}{\sum_{\mp_i \in \Omega(\mx)} \theta(\|\mx - \mp_i\|, r_i)},
\end{equation}
where $\Omega(\mx)$ denotes the set of MLS points that are inside the ball centered at $\mx$ with radius $r_b$. $r_b$ can be set by the user as the truncating point distance values.

Due to the above formulation, the zero function values $F(\mx)$ exist in a narrow band region of the IMLS points. Extracting  $\mS$ explicitly as a triangle mesh can be done efficiently via Marching cubes~\cite{marchingcubes}, by restricting the functional evaluation on the regular grids inside the bounded region.

\subsection{Deep IMLS surface} \label{subsec:dimls}
Sparse and unoriented point clouds probably with noise and missing regions are typical inputs for 3D reconstruction in practice, but they cannot be handled well by traditional 3D reconstruction methods like Poisson reconstruction. To deal with this kind of imperfect data,  we aim to design an auto-encoding neural network to generate IMLS surfaces.

A na\"{i}ve way of defining the network output is to set a fixed-number of IMLS point tuples: $\{\mp_i, \mn_i, r_i\}_{i=1}^N$, similar to existing point set generation approaches~\cite{Su2017}. However, it will constrain the representation capability of IMLS and cannot learn local geometry priors well from the data.  We introduce an intermediate network output -- \emph{octree-based scaffold}, to help generate MLS points as needed. The octree-based scaffold is a $d$-depth octree $\mO$ which roughly approximates the 3D surface in multi-resolutions. For each finest non-empty octant $o_k$, \ie the smallest non-empty voxel in the octree,  we associate a small set of MLS points whose locations are near to the octant center $\mc_k$. To be specific, the MLS points associated with $o_k$ are defined as
\begin{equation}
    \mp_{k,l} :=\mc_k + \mt_{k,l}, \quad l=1,\ldots,s,
\end{equation}
where $\mt_{k,l} \in \mathbb{R}^3$ is the offset vector from $\mp_{k,l}$ to $\mc_k$, $s$ is a predefined point number.
As the structure of the octree-based scaffold depends on the target surface, the total number of MLS points and their positions are determined adaptively.

With the setup above, a suitable network for IMLS generation should output: (1) an octree-based scaffold $\mO$; (2) the MLS point offsets, MLS point normals and control radii for each finest non-empty octant $o_k$, denoted by  $\mt_{k,l}, \mn_{k,l}$ and $r_{k,l}$, respectively.

Here we note that the octree created from the input noisy and sparse point cloud cannot be used as the scaffold, as it could be incomplete and inaccurate, and different from that of the target shape.

\myparagraph{Scaffold prediction}
We use the octree-based convolutional neural network (O-CNN) autoencoder~\cite{Wang2017,Wang2018a} to generate the scaffold.  Its encoder takes a $d_{in}$-depth octree as input which is constructed from the input point cloud,  and performs the CNN computation within the octree only. Its decoder starts with $4\times4\times4$ cells and predicts whether each cell is empty or not, and subdivides it into eight octants if the cell is not empty. This process is performed on each non-empty octant recursively until the maximum output octree depth $d_{out}$ is reached.

\myparagraph{MLS point prediction}
Unlike previous works \cite{Wang2018a,Wang2020} where the decoders regress an oriented point or a planar patch at each finest non-empty octant for achieving sub-voxel precision, we predict $s$ MLS point tuples based on the feature vector at the octant, denoted by $f(o_k)$, via a
multilayer perceptron (MLP) with one hidden-layer as follows:
\begin{equation}
    \texttt{MLP} \circ f(o_k) = (\mt_{k,1}, \mn_{k,1}, r_{k,1}, \cdots, \mt_{k,s}, \mn_{k,s}, r_{k,s}).
\end{equation}
Note that the MLP predicts the local coordinates of $\mp_{k,l}$, \ie $\mt_{k,l}$, thus it can learn the local prior from the data.  To ensure $\mp_{k,l}$ is close to $\mc_k$, the value range of each coordinate component of $\mt_{k,l}$ is restricted within $[-\beta  h, \beta  h]$, here $h$ is the size of the finest octant and $\beta$ is set to $1.5$ by default. We also constrain $r_{k,s}$ within $[l_r/2,2l_r]$, where $l_r=h/\sqrt{s}$.  These constraints are implemented by using the $\tanh$ activation for the network output and scaling the value by its range scale.

\subsection{Network structure} \label{subsec:net}

We use a U-Net-like O-CNN autoencoder~\cite{Wang2020} that contains O-CNN ResNet blocks and output-guided skip-connections for better preserving the input geometry and predicting missing regions. The output-guided skip-connection is the skip connection added between each non-empty octant of the output octree and its corresponding octant of the input octree at the same octree level, if the latter octant exists.

For a given unoriented point cloud, we construct a $d_{in}$-depth octree for it. At each finest octant, we set the input 4-dimensional signal by concatenating the offset from the average position of input points inside the octant to the octant center with a binary scalar indicating whether the octant is empty.

\begin{figure*}[t]
    \centering
    \begin{overpic}[width=0.9\linewidth]{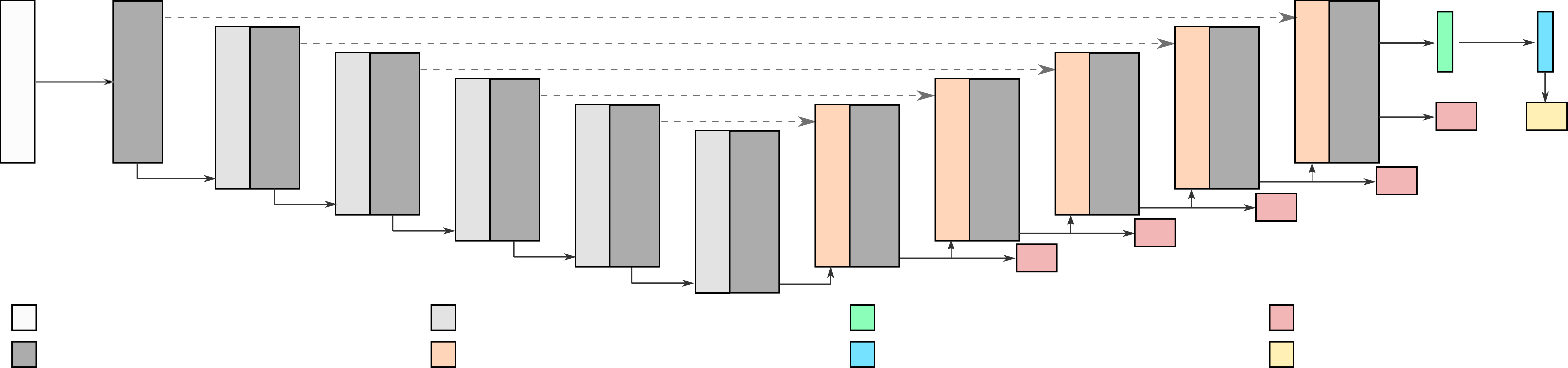}
        \put(3,2.6){\small Input}
        \put(3,0.3){\small $\mathrm{Resblock}(n,c)$}
        \put(30,2.6){\small $\mathrm{Downsample}(c)$}
        \put(30,0.3){\small $\mathrm{Upsample}(c)$}
        \put(57,2.6){\small MLP for IMLS prediction}
        \put(57,0.3){\small IMLS parameters}
        \put(84,2.6){\small Octree structure loss}
        \put(84,0.3){\small IMLS loss}
    \end{overpic}
    \caption{Deep IMLS Network structure for 3D reconstruction from point clouds. Each block after the input represents the CNN computation at the octants in one level of the octree, from fine to coarse on the encoder, and vice versa on the decoder. } \label{fig:net} 
\end{figure*}
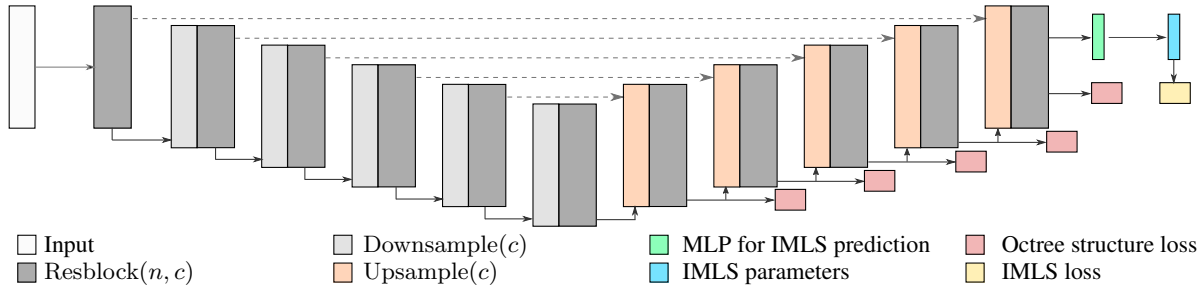

The network structure is illustrated in \cref{fig:net}. $\mathrm{Resblock}(n,c)$ denotes an $n$-layer O-CNN-based residual block with channel number $c$. $\mathrm{Downsample}(c)$ and $\mathrm{Upsample}(c)$ are octree-based convolution and deconvolution operators ~\cite{Wang2017} followed by batch normalization and ReLU. $c$ is set to 64 for the first Resblock, and increases by a factor of 2 after each Downsample operator, and divided by 2 after each Upsample operator. In our experiments, we set $n=3$.  A hidden-layer MLP is used for predicting whether the octant is empty or not.

\subsection{Loss function design} \label{subsec:loss}
Our network is trained in a supervised manner. For each shape in the training dataset, we precompute its ground-truth $d_{out}$ octree $\mO_o$ and use it for supervising the scaffold prediction. The octants at the first and second levels of $\mO_o$ are set to be non-empty.
A succinct set $\mQ$ of sample points which bound the ground truth surface tightly will be used for probing the implicit function accurately. For generating these points, please refer to Section \ref{para:dataset}.
We then compute the ground-truth signed distance function (SDF) values $F_o(\mq)$ and the SDF gradient $\nabla F_o(\mq)$, for each $\mq\in \mQ$. Our main training objective is to fit the signed distance field sampled on $\mQ$.

The loss function consists of the following terms.

\myparagraph{Octree structure loss}
The determination of octant status is a binary classification problem: 0 for empty and 1 for non-empty. We use the weighted summation of the sigmoid cross-entropy loss at every octant of $\mO$ to define the octree structure loss.
\begin{equation}
    L_{\texttt{oct}} := \lambda_{o} \sum_{j=3}^{d_{out}} \frac{1}{|\mO_j|} \sum_{o \in \mO_j} \mathrm{cross\_entropy}(o),
\end{equation}
here $\mO_j$ is the octant set at level $j$, and $|\mO_j|$ denotes its size.

\myparagraph{SDF loss}  The difference between the predicted IMLS surface and the ground-truth SDF field is defined by the SDF loss $L_{\texttt{sdf}}$.
\begin{equation}
    L_{\texttt{sdf}} := \sum_{\mq \in \mQ} \lambda_s  \|F(\mq) - F_o (\mq) \|^2 +  \lambda_g  | 1 - \nabla F(\mq)  \cdot \nabla F_o (\mq) |.
\end{equation}

Here the gradient of $F$ can be approximated by:
\begin{equation}\label{eq:mls_grad}
    \nabla F(\mx) \approx  \dfrac{\sum_{\mp_i \in \Omega(\mx,\mP)} \theta(\|\mp_i-\mx\|, r_i) \cdot  \mn_i}{\sum_{\mp_i \in \Omega(\mx,\mP)} \theta(\|\mp_i-\mx\|, r_i)}.
\end{equation}
To speed up the computation, we restrict the region of $\Omega(\mx,\mP)$ by the octants whose distances from their centers to $\mx$ are less than $4h$.

\myparagraph{MLS point repulsion loss} Inspired by the repulsion term used by \cite{Huang2009} for distributing consolidated points regularly, we introduce the point repulsion loss to improve the local regularity of our generated MLS points.

\begin{equation}
    L_{\texttt{rep}} := \lambda_{rep}  \sum_{\mp_i} \sum_{\mp_j \in \Omega(\mp_i)} -w_{ij}  \|\mp_i-\mp_j\|_{\texttt{proj}},
\end{equation}
where $\|\mp_i-\mp_j\|_{\texttt{proj}} = \|(\mathbf{I} - \mn_{i}\mn_{i}^T)(\mp_i-\mp_j)\|$ is the length of the projection of $\mp_i-\mp_j$ onto the tangent plane at $\mp_i$, and $w_{ij}$ is a bilateral weight with respect to both the MLS point difference and normal difference, defined as follows.
\begin{equation}
    w_{ij} = \exp\bigl(-\|\mp_i -\mp_j \|^2 /r_j^2 - (1 - \left< \mn_i, \mn_j \right>)\bigr).
\end{equation}
The above design pushes $\mp_j$ away from $\mp_i$, especially when their normals and their positions are similar.

\myparagraph{Projection smoothness loss} For achieving local surface smoothness, we encourage MLS points to be close to the tangent planes of their neighboring MLS points.
\begin{equation}\label{eq:proj}
    L_{\texttt{proj}} = \lambda_{p} \sum_{\mp_i} \sum_{\mp_j \in \Omega(\mp_i)} w_{ij}  \left<\mn_i, \mp_i -\mp_j \right>^2. \\
\end{equation}

\myparagraph{Radius smoothness loss}
Similarly, for improving surface smoothness, the radius change of neighboring MLS points is penalized via weighted-Laplacian smoothing on radii.
\begin{equation}\label{eq:radius}
    L_{\texttt{rad}} = \lambda_{r} \sum_{\mp_i} \bigl\| r_i - \dfrac{\sum_{\mp_j \in \Omega(\mp_i)} w_{ij} \cdot r_j}{\sum_{\mp_j \in \Omega(\mp_i)} w_{ij}} \bigr\|^2.
\end{equation}

\myparagraph{Weight decay} A small weight decay with a coefficient $\lambda_w$ is added to the loss function.

The above terms are added up together to form the total loss function. The coefficients of loss terms are specified in \cref{sec:result}.

\section{Experiments} \label{sec:result}

We evaluate the efficacy of IMLSNets for 3D object reconstruction and compare it against the state-of-the-art 3D reconstruction methods. Extensive ablation tests are reported in \cref{subsec:ablation}.

\myparagraph{Dataset}\label{para:dataset}
We use 3D models from 13 shape classes of ShapeNet~\cite{shapenet2015} as our training and test datasets. The dataset split strategy and the input point cloud preparation follow the setup of \cite{Mescheder2019,Peng2020}: each input point cloud contains $3000$ points, randomly sampled from the ground-truth shape and Gaussian noise with standard deviation set to 0.005 of the maximum bounding box side length of the shape. For each 3D model, we prepare its ground-truth octree for training.
The ground-truth octree is built, by recursively subdividing the nonempty octants from the root node to the maximum depth $d_{out}$ \cite{Wang2017}.  The set $\mathbf{Q}$ of grid points slightly larger than the nonempty leaf nodes of the octree is built to probe the SDF values for implicit function supervision. The exact training data preparation is provided in \cref{appendix:data}. 

\myparagraph{Parameters}
The octree depth for storing the ground-truth signed distance field is 7. The coefficients of loss terms are set as follows: $\lambda_o = 0.1,\lambda_g = 0.05$, $\lambda_p = \lambda_r=10$, $\lambda_{rep}=0.05$,  $\lambda_w = 5\times 10^{-5}$. The coefficient of the SDF loss term depends on the octree level of SDF sample points: $\lambda_s = 200$ if the level is $6$, $\lambda_s = 800$ if the level is $7$.  The selection of neighboring MLS points $\Omega(\mx)$ is speedup via space partitions:  $10$ nearest MLS points are selected in our experiments.

\myparagraph{Network configuration}
We use the octree-based encoder and decoder with output-guided skip connections~\cite{Wang2020}. The network is denoted as IMLSNet($d_{in}$,$d_{out}$,$s$), where $d_{in}$ and $d_{out}$ are the input and output octree depths, and $s$ is the number of MLS points associated with each non-empty finest octants. IMLSNet(7,7,1) is set by default.

\myparagraph{Network training}
Our networks were trained using Adam optimizer (lr=0.001), with batch size 32.  We decay the learning rate by 20\% after every 10 epochs if it is greater than 0.0001. The gradient update of $w_{ij}$ is disabled in our training as we found that back-propagating through $w_{ij}$ to position and normal variables causes training instability, especially during the early stages when network predictions are very inaccurate. We follow the curriculum SDF training strategy~\cite{Duan2020} to improve the training quality: the SDF sample points at the coarse octree level (depth=6) are first used, and after 30 epochs, we use sample points at a deeper level (depth=7).  All the experiments were run on a Linux server with an Intel Core I7-6850K CPU (3.6GHz) and a GeForce GTX 1080 Ti GPU (11 GB memory). The network parameter size of IMLSNet(7,7,1) is about 4.3 M.
More details are reported in \cref{appendix:net}, and our code is available at \url{https://github.com/Andy97/DeepMLS/}.

\myparagraph{Network efficiency} Averagely, our network forwarding time takes 30 milliseconds per point cloud input. The generated MLS points present a good preview of the final surface. Our na\"{i}ve CPU-based marching-cube implementation takes 1 second averagely for a single shape, including file IO time.

\myparagraph{Evaluation metrics} We evaluate the reconstruction quality of our results by a set of common metrics: $L_1$-Chamfer distance ($\texttt{CD}_1$) and normal difference (ND) measure bi-directional point-wise and normal error from ground truth; F-score($\tau$)~\cite{Tatarchenko2019} is the harmonic mean of recall and precision under the tolerance $\tau$;
Occupancy IOU (IOU) computes the volumetric IOU. Our surface results are triangle meshes extracted via Marching cubes on a $128^3$ grid. The CD metric is scaled by $10$, $\tau$ is set to 0.1 by default.
Similar metrics on the MLS point sets are also measured to reflect the reconstruction capability of MLS point sets.  The detailed metric formulae and other measurements are provided in \cref{appendix:metric}.

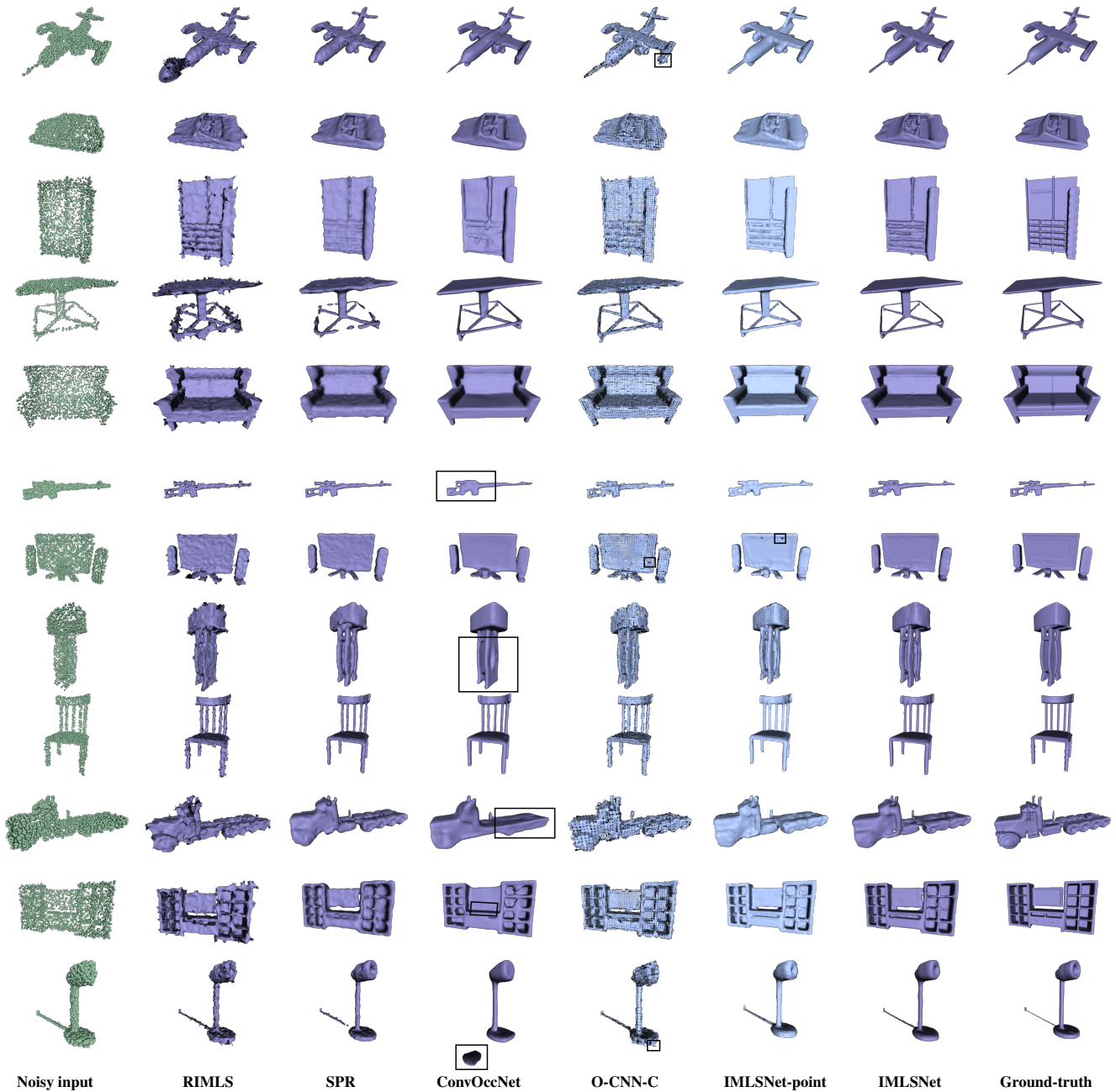
\begin{figure*}[t]
    \centering
    \begin{overpic}[width=1\linewidth]{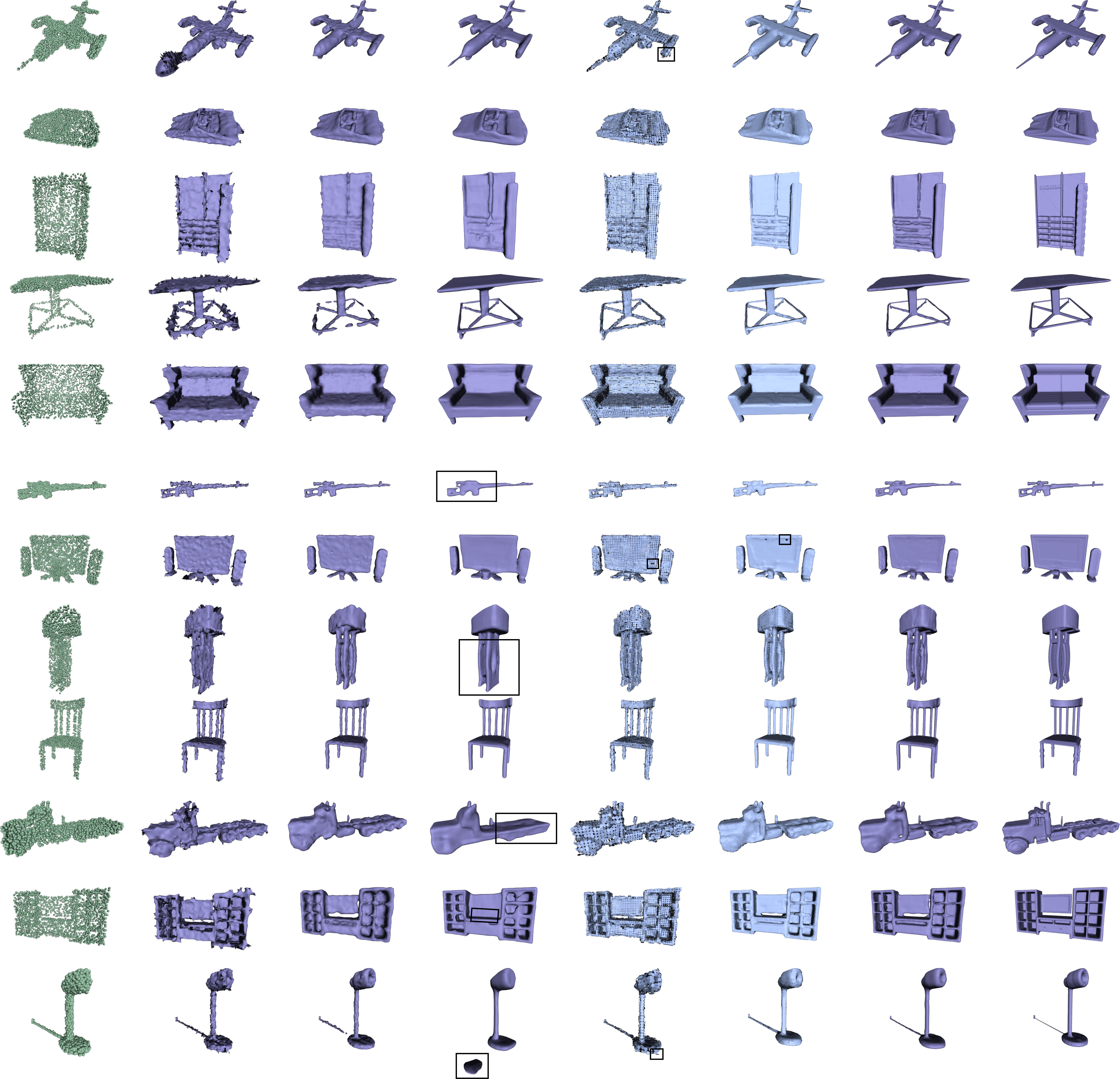}
        \put(1,-1.5){\scriptsize  \textbf{Noisy input}}
        \put(16,-1.5){\scriptsize  \textbf{RIMLS}}
        \put(29,-1.5){\scriptsize \textbf{SPR}}
        \put(39,-1.5){\scriptsize  \textbf{ConvOccNet}}
        \put(53,-1.5){\scriptsize  \textbf{O-CNN-C}}
        \put(65,-1.5){\scriptsize  \textbf{IMLSNet-point}}
        \put(79,-1.5){\scriptsize \textbf{IMLSNet}}
        \put(90,-1.5){\scriptsize  \textbf{Ground-truth}}
    \end{overpic}
    \vspace{1mm}
    \caption{Visual results of 3D object reconstruction from point clouds by different methods. For point outputs of O-CNN-C and IMLSNet-point, we render points as oriented disks. Some artifacts in the results are labeled by black boxes. }
    \label{fig:shape} \vspace{-2mm}
\end{figure*}

\subsection{3D object reconstruction from point clouds} \label{subsec:shape_result}
As IMLSNets learn local priors, the learned features do not rely on specific shape classes. We trained IMLSNet(7,7,1) on the ShapeNet dataset (13 classes) in a class-agnostic way and tested our network on their test data of 13 classes, and 5 unseen shape classes (bathtub, bag, bed, bottle, pillow). We also compared our method with two classical reconstruction methods: robust implicit moving least squares (RIMLS)~\cite{Oztireli2009} and Screened Poisson reconstruction (SPR); and three learning-based methods: occupancy network (OccNet)~\cite{Mescheder2019},
convolutional occupancy network(ConvOccNet)~\cite{Peng2020} and O-CNN based completion network (O-CNN-C)~\cite{Wang2020}. OccNet and ConvOccNet are state-of-the-art deep implicit function-based methods and the work of ~\cite{Peng2020} shows that ConvOccNet performs better than both OccNet and DeepSDF~\cite{Park2019}. O-CNN-C generates a set of oriented points on the predicted octree. For RIMLS, Screened Poisson, and O-CNN-C, we provide the ground-truth point normals as the input. We use the pretrained O-CNN-C whose depth of the output octree is 6.

\cref{tab:shape} shows that our network achieves the best performances among all the compared methods. From the visual comparison (\cref{fig:shape}), we can observe that all the learning-based methods handle noisy and incomplete point inputs much better than classical non-learning reconstruction methods. Our IMLSNet results contain more details and capture small parts better, while other deep implicit-function methods tend to miss these details (see the examples in the 6th, 8th and 10th rows).  This advantage of IMLSNet over ConvOccNet which uses fixed grids and implicit functions is due to the proper combination of adaptive and sparse grids (octree), point set, and implicit representations that together allow for more adaptive modeling of detailed geometry. In \cref{tab:lfd}, the perception quality comparison measured by Light Field Descriptor (LFD)~\cite{Chen2019} on five categories further confirms the superiority of our approach.

Our MLS point sets (IMLSNet points) are also good discrete approximations and perform better than O-CNN-C whose results contain small missing regions and floating points (see the examples in the first and last rows). The small missing region in IMLSNet points can be recovered by the final IMLS surface.  More visual comparisons can be found in \cref{appendix:more}.

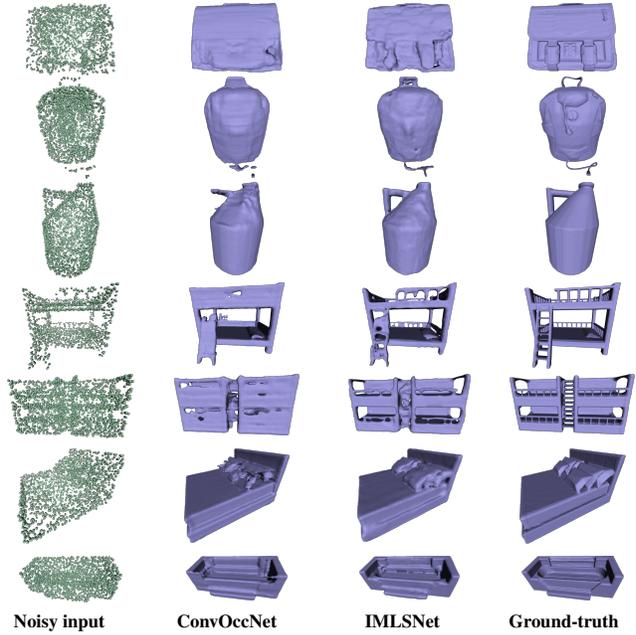
\begin{figure}[t]
    \centering
    \begin{overpic}[width=1\linewidth]{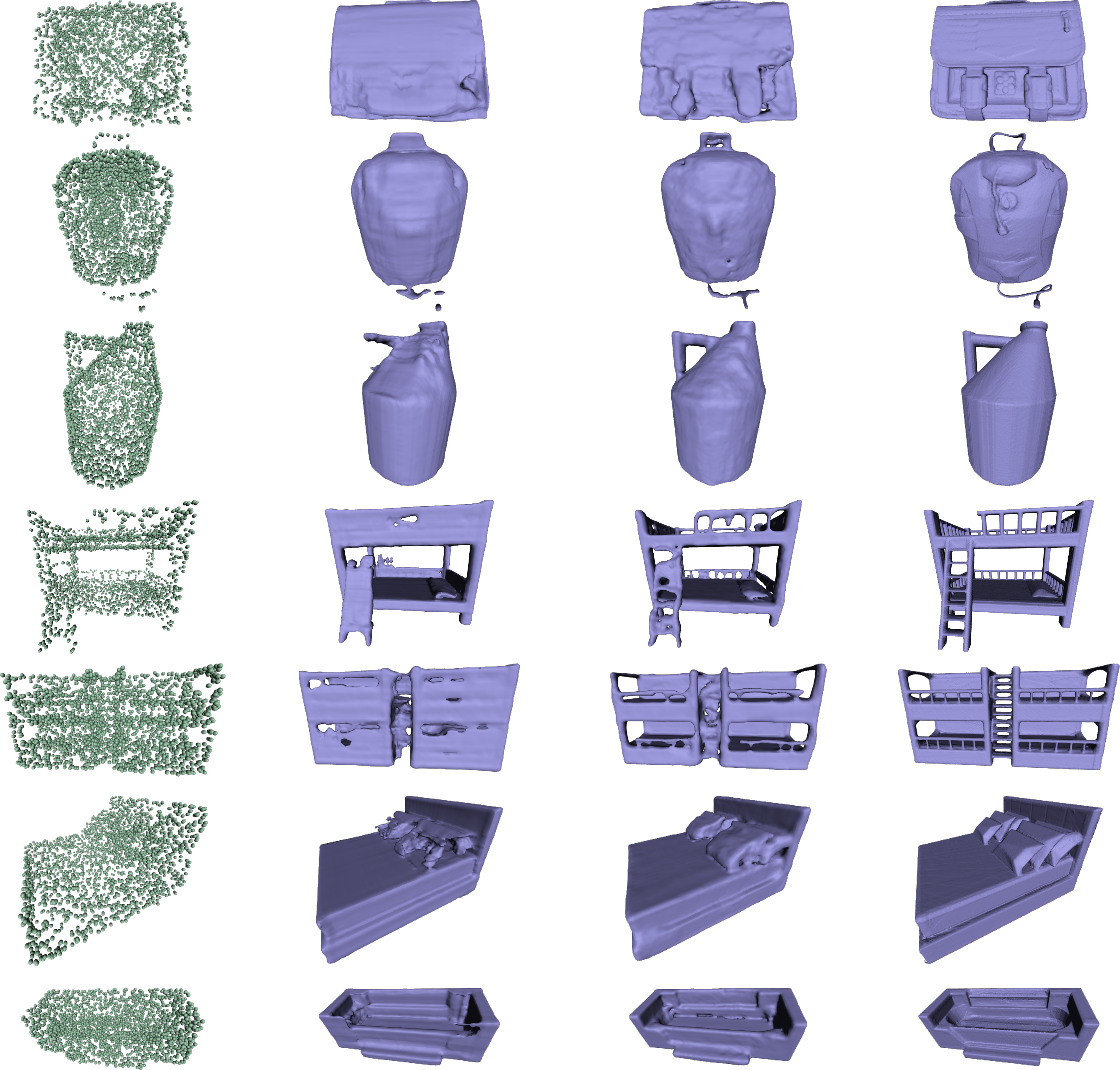}
        \put(1,-4){\scriptsize \textbf{Noisy input}}
        \put(27,-4){\scriptsize  \textbf{ConvOccNet}}
        \put(57,-4){\scriptsize  \textbf{IMLSNet}}
        \put(80,-4){\scriptsize  \textbf{Ground-truth}}
    \end{overpic}
    \vspace{1mm}
    \caption{Visual results of 3D object reconstruction on five unseen shape classes. From left to right: the input point, ConvOccNet, our IMLSNet, and the ground-truth model.}
    \label{fig:shapeunseen} 
\end{figure}

The statistics and illustrations on the unseen shape classes (\cref{fig:shapeunseen}) also confirm the generalization ability of all the approaches that use local features. Among them, our method achieves the best result.

\begin{table}[t]
  \centering
  \scalebox{0.74}{
    \begin{tabular}{cllll}
      \toprule
      \thead{ Network } & \thead{$\texttt{CD}_1$  $\downarrow$  } & \thead{NC  $\uparrow$}        & \thead{IoU $\uparrow$}        & \thead{F-Score $\uparrow$}    \\
      \midrule
      O-CNN-C            & 0.067(0.067)                             & 0.932(0.945)                   & n/a                            & 0.800(0.802)                   \\
      IMLSNet pts.       & 0.035(0.037)                             & 0.941(0.954)                   & n/a                            & 0.985(0.983)                   \\
      \midrule
      OccNet             & 0.087                                    & 0.891                          & 0.761                          & 0.785                          \\
      ConvOccNet         & 0.044(0.053)                             & 0.938(0.948)                   & 0.884(0.902)                   & 0.942(0.916)                   \\
      IMLSNet            & \textbf{0.031}(\textbf{0.034})           & \textbf{0.944}(\textbf{0.956}) & \textbf{0.914}(\textbf{0.939}) & \textbf{0.983}(\textbf{0.981)} \\
      \bottomrule
    \end{tabular}
  }
  \caption{Quantitative evaluation of different networks on the test data of 13 shape classes and the full data of 5 unseen shape classes. Numbers in parentheses are for the unseen classes and numbers in bold are the best.  The evaluation on each shape class and the training configuration of compared networks are reported in \cref{appendix:more}. }
  \label{tab:shape}
\end{table}

\begin{table}[t]
    \centering
    \scalebox{0.82}{
        \begin{tabular}{cccccc}
            \toprule
            \textbf{Network}  & \textbf{plane }       &   \textbf{car}    & \textbf{chair}  & \textbf{rifle} & \textbf{table}\\
            \midrule
            ConvOccNet & 1540.47 & 731.84 & 908.04 &   1433.82 &  660.10 \\
            IMLSNet & \textbf{1326.22} &   \textbf{649.69} &\textbf{710.38} & \textbf{1211.96} &  \textbf{608.87} \\
            \bottomrule
        \end{tabular}
    }
    \caption{LFD metric evaluation (lower values are better). The comparison between IMLSNet(7,7,1) and ConvOccNet shows that the visual quality of IMLSNet is better.}
    \label{tab:lfd}
    \vspace{-2mm}
\end{table}

\subsection{Ablation study} \label{subsec:ablation}
A series of experiments on 3D object reconstruction from point clouds were conducted to evaluate each component and parameter choice of our network. The quantitative comparison on the test dataset is reported in \cref{tab:ablation}.

\begin{figure}[t]
    \centering
    \begin{overpic}[width=\linewidth]{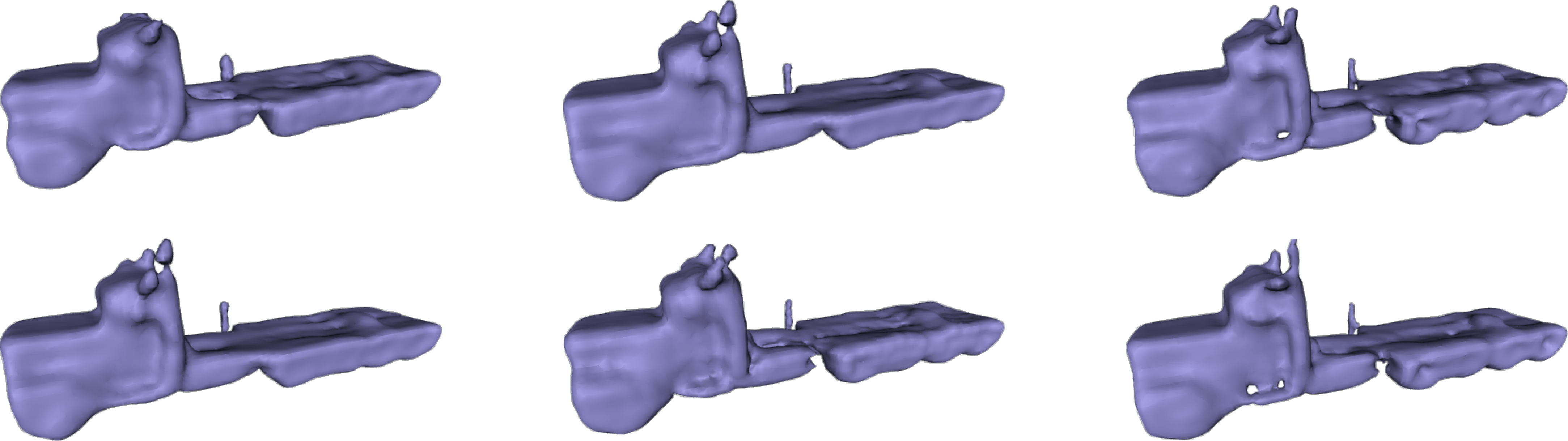}
        \put(8,14){\scriptsize \textbf{IMLSNet(7,5,1)}}
        \put(42,14){\scriptsize \textbf{IMLSNet(7,6,1)}}
        \put(78,14){\scriptsize \textbf{IMLSNet(7,7,1)}}
        \put(8,-3){\scriptsize \textbf{IMLSNet(6,6,1)}}
        \put(42,-3){\scriptsize \textbf{IMLSNet(7,5,4)}}
        \put(78,-3){\scriptsize \textbf{IMLSNet(7,6,2)}}
    \end{overpic}
    \vspace{0.2mm}
    \caption{Ablation study of IMLSNets under different octree depths and different MLS point numbers in an octant. }
    \label{fig:depth} \vspace{-1mm}
\end{figure}
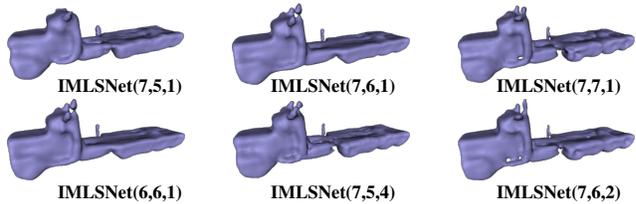

\myparagraph{Input and output depth} By changing the depth number of the input and output octrees, our experiments reveal that: (1) the network with the deeper input octree like IMLSNet(7,6,1) yields more accurate results than IMLSNet(6,6,1) as more input point information are fed to the network; (2) the network with the deeper output octree like IMLSNet(7,7,1), also yields better-detailed geometry than the networks with shallow output octrees, like IMLSNet(7,6,1) and IMLSNet(7,5,1). These experiments show that the octree input and the octree-based scaffold are helpful to gather more point information and predict shape geometry. \cref{fig:depth} illustrates their visual differences on the reconstruction results of a noisy truck model.

\myparagraph{Number of MLS points} We also found that using shallow output octree with a larger $p$, like IMLSNet(7,6,2) and IMLSNet(7,5,4), can achieve similar results by a deeper output octree with a smaller $p$, like IMLSNet(7,7,1), as the total number of MLS points of these networks are comparable. However,  we observe IMLSNet(7,7,1) is better at capturing small and thin details (see \cref{fig:depth}).

\begin{figure}[t]
    \centering
    \begin{overpic}[width=\linewidth]{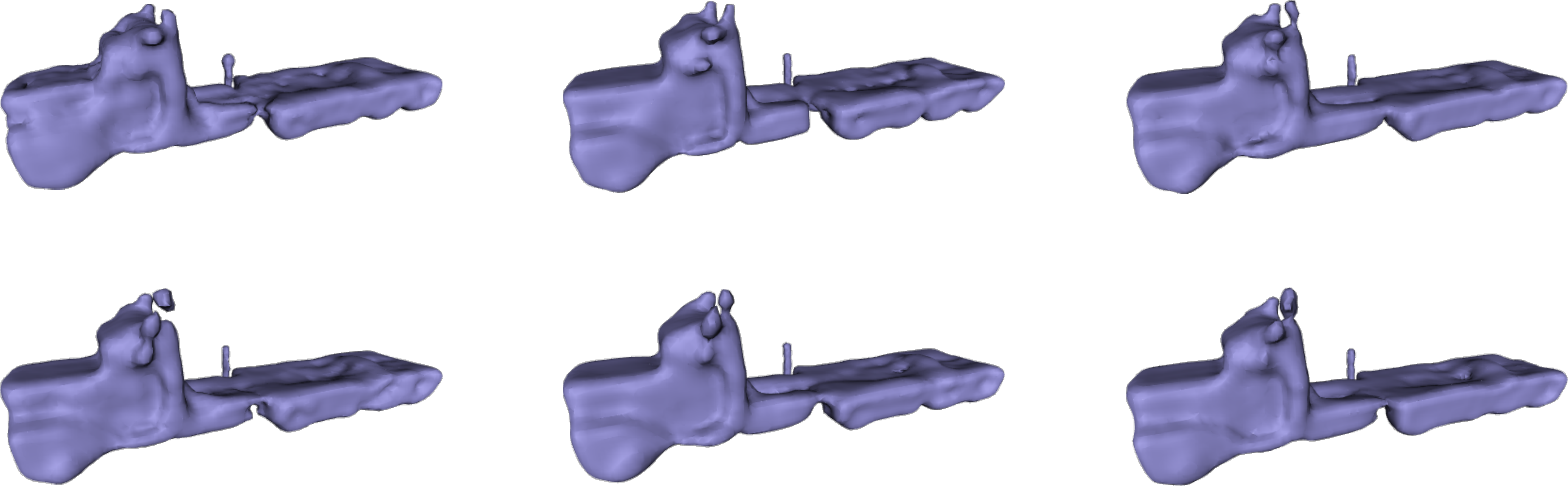}
        \put(1,16){\scriptsize \textbf{cr.(6,6,1)}}
        \put(38,16){\scriptsize \textbf{cr.(7,7,1)}}
        \put(73,16){\scriptsize \textbf{w/o ps.(6,6,1)}}
        \put(1,-3){\scriptsize \textbf{w/o rep.(6,6,1)}}
        \put(37.5,-3){\scriptsize \textbf{w/o rs.(6,6,1)}}
        \put(75,-3){\scriptsize \textbf{w/o grad.(6,6,1)}}
    \end{overpic}
    \vspace{0.2mm}
    \caption{Visual results of IMLSNets under different configurations for a truck model. }
    \label{fig:others2} \vspace{-4mm}
\end{figure}
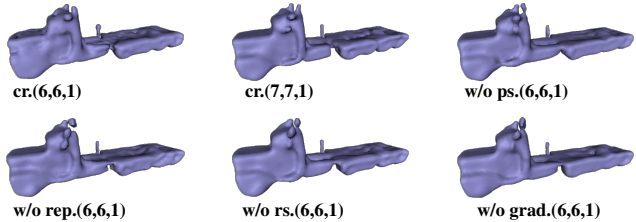

\myparagraph{Ablation study of loss terms} By dropping the surface gradient term, the repulsion term, the projection smoothness term, and the radius smoothness term independently from IMLSNet(6,6,1), our experiments show that the surface quality degrades accordingly (see \cref{tab:ablation} and \cref{fig:others2}). The tested networks are denoted by \emph{w/o grad.(6,6,1)}, \emph{w/o rep.(6,6,1)}, \emph{w/o ps.(6,6,1)}, \emph{w/o rs.(6,6,1)}. The gradient term plays a more important role than other terms, as the network trained without this term has the lower IoU and NC, also loses some surface details. The radii smoothness term is less important and the performance of the corresponding network slightly drops compared to IMLSNet(6,6,1).  More visual comparisons are provided in \cref{appendix:more}.

\myparagraph{Constant MLS radius} We fixed the MLS point radius to $l_r$ and trained two networks: IMLSNet(6,6,1) and IMLSNet(7,7,1), denoted by \emph{cr.(6,6,1)} and \emph{cr.(7,7,1)}. The results (see \cref{tab:ablation} and \cref{fig:depth}) show that the former network yields worse results than the default IMLSNet(6,6,1), but the latter network performs well. This phenomenon indicates that using the adaptive MLS radii would be more effective for IMLSNets with coarser scaffolds.

\begin{table}[t]
  \centering
  \scalebox{0.9}{
  \begin{tabular}{rcccc}
  \toprule
 \thead{ Network }            & \thead{$\texttt{CD}_1$  $\downarrow$  }       &   \thead{NC  $\uparrow$}    & \thead{IoU $\uparrow$} & \thead{F-Score $\uparrow$}\\
  \midrule
  IMLSNet(7,5,1)      &  0.038 &  0.928 &  0.888 &  0.962  \\
  IMLSNet(7,6,1)      &  0.031 &  0.943 &  0.913 &  0.981  \\
  IMLSNet(6,6,1)      &  0.033 &  0.941 &  0.904 &  0.977  \\
  IMLSNet(7,7,1)      &  0.031 &  \textbf{0.944} &  0.913 &  0.983  \\
  \midrule
  IMLSNet(7,5,4)      &  0.032 &  0.936 &  0.905 &  0.977  \\
  IMLSNet(7,6,2)      &  \textbf{0.030} &  \textbf{0.944} &  \textbf{0.915} &  0.983  \\
  \midrule
  w/o grad.(6,6,1)     &  0.034 &  0.938 &  0.897 &  0.974  \\
  w/o rep.(6,6,1)    &  0.034 &  0.939 &  0.899 &  0.976  \\
  w/o ps.(6,6,1)       &  0.034  &  0.939 &  0.900 &  0.974  \\
  w/o rs.(6,6,1)    &  0.033  &  0.940 &  0.902 &  0.977  \\
  \midrule
  cr.(7,7,1) &  0.031 &  \textbf{0.944} &  0.913 &  \textbf{0.984}  \\
  cr.(6,6,1) &  0.036 &  0.933 &  0.880 &  0.966  \\
   \bottomrule
  \end{tabular}
  }
  \caption{Quantitative evaluation of IMLSNets under different configurations.
  }
  \label{tab:ablation}
  \end{table}

\begin{figure}[t]
  \centering
  \includegraphics[width=0.9\linewidth]{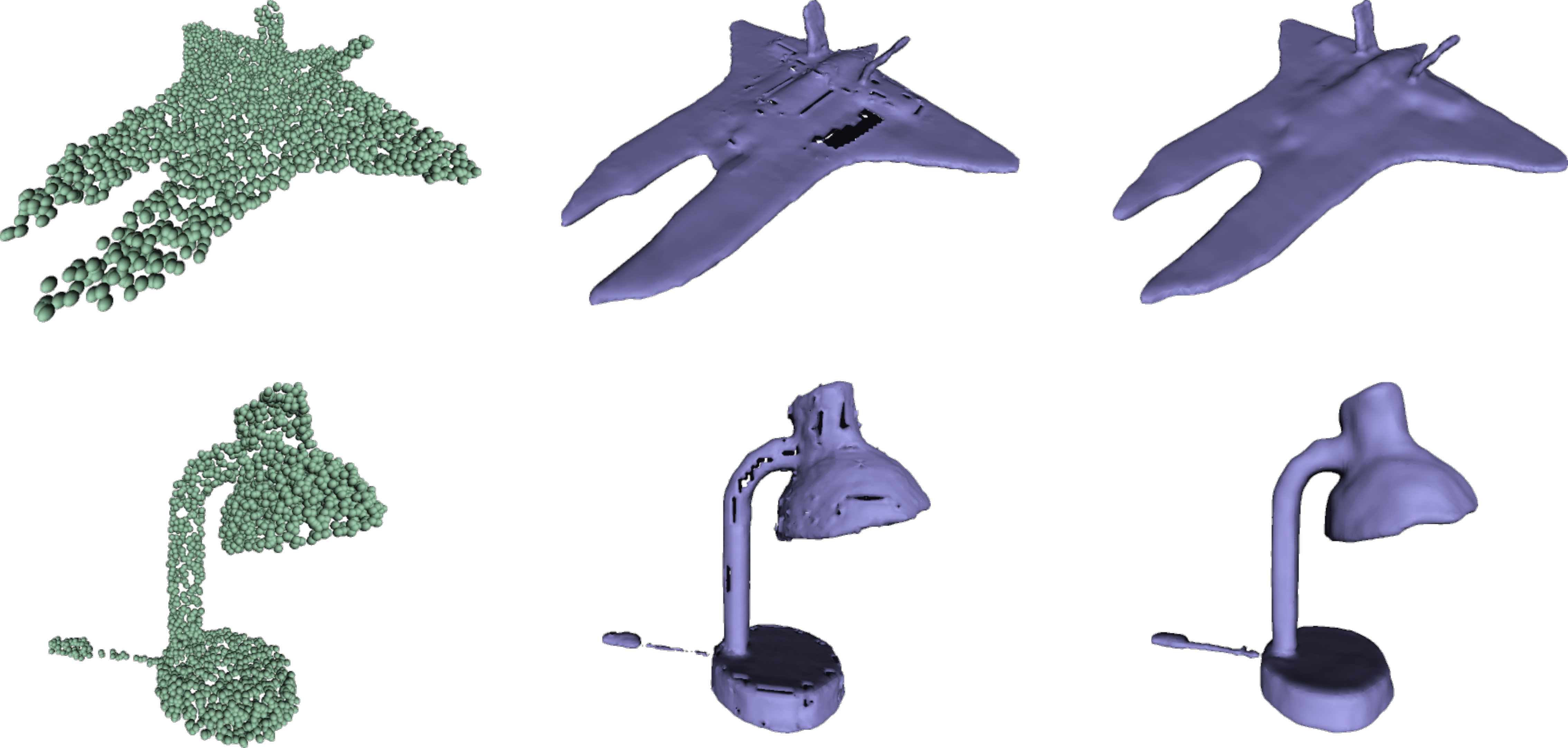}
  \caption{The results of octree-aided deep local implicit function (middle) and our IMLS surface results (right) from the same noisy inputs (left).}
  \label{fig:local} \vspace{-2mm}
\end{figure}

\myparagraph{Octree-aided deep local implicit function} Instead of predicting MLS points at each non-empty octant, one can follow \cite{Peng2020} to predict the local implicit function value from the interpolated feature vector in the octant via MLP directly.  We call this approach \emph{Octree-aided deep local implicit function}. We implemented this approach and compared it with our IMLSNet. Our experiments (using depth-6 octrees) show that this approach can recover most surface regions with good quality, but the results may contain some holes as there is no guarantee that the zero-iso surface always passes through the finest non-empty octant region,  while our IMLSNet can do not have this flaw due to the explicit point generation and IMLS surface formulation. \cref{fig:local} illustrates this phenomenon on a plane model and a lamp model. More illustrations and discussions are presented in \cref{appendix:discussion}.

\section{Conclusion and future work} \label{sec:conclusion}
We present a deep implicit moving least-squares surface technique for 3D reconstruction, which enjoys the flexibility of point sets and the approximation power of implicit surface representations.  Its efficacy and generalization ability are well demonstrated through extensive tests.

\myparagraph{Beyond planes} A straightforward extension for enhancing reconstruction quality is to use higher-order IMLS like APSS~\cite{Guennebaud2007,Guennebaud2008} which replace planes with spheres,  and other variants of MLS surfaces.

\myparagraph{Differentiable rendering} As the band regions of MLS surfaces can be determined by MLS points, rendering MLS surfaces by sphere marching algorithms can be performed more efficiently than other non-point-based implicit representations. Combining IMLS-Net and implicit surface ray-tracing will enable differentiable MLS prediction with 2D image supervision only.

\myparagraph{MLS point generation} As less MLS points can well approximate relatively planar and simple surface regions, devising an adaptive and non-uniform MLS point generation scheme would help reduce unnecessary MLS points and increase IMLS reconstruction quality further. Using and predicting the adaptive octrees~\cite{Wang2018a} as the scaffold could be a promising solution.

{\small
  \bibliographystyle{ieee_fullname}
  \bibliography{src/reference}
}
\newpage
\appendix

\section{Training data preparation} \label{appendix:data}
For training data involved with the ShapeNet dataset, we use data preprocessing tools from \cite{Peng2020} to generate watertight meshes via TSDF fusion. We then normalize each mesh into a $[-1, 1]^3$ bounding box with 5\% padding and compute signed distance function (SDF) values and  gradients using the OpenVDB library (\url{https://www.openvdb.org}). We generate $256 \times 256 \times 256$ SDF grids, denoted by $\mathcal{F}=\{(i,j,k,s_{i,j,k}, \nabla s_{i,j,k})\}$, and collect SDF samples subset in a progressive manner: we first gather depth-6 SDF samples ((\ie samples whose indices satisfy: $i,j,k  \mod 4  = 0$)) with absolute SDF values less than $\frac{1}{8}$, this threshold guarantees coverage of generated octree nodes. To better capture shape details, similar to the sampling strategy in \cite{Park2019}, we add more SDF samples near the surface, to be concrete, the depth-7 SDF samples with absolute SDF values less than $\frac{1}{16}$.

\section{Evaluation metrics} \label{appendix:metric}
We reuse the evaluation tools of \cite{Peng2020} to compute the following metrics. We denote $M_{g}$ and $M_{p}$ as the ground-truth mesh and the mesh of the predicted result. $\mathcal{X}:=\{\mx_1, \ldots, \mx_{N_g}\}$ and $\mathcal{Y}: =\{\my_1, \ldots, \my_{N_p}\}$ are randomly sample points on these two meshes, respectively.  We define $\proj_{g2p}(\mx) = \argmin_{\my \in \mathcal{Y}} \| \mx - \my \|$ and $\proj_{p2g}(\my) = \argmin_{\mx \in \mathcal{X}} \| \mx - \my \|$. $\mn(\cdot)$ denote an operator that returns the normal vector of a given point.
\begin{itemize}
    \item $L_1$ Chamfer distance.
          \begin{align*}
              \texttt{CD}_1 = & \frac{1}{2N_g}\sum_{i=1}^{N_g} \| \mx_i - \proj\nolimits_{g2p}(\mx_i) \| + \\ & \frac{1}{2N_p}\sum_{i=1}^{N_p} \| \my_i - \proj\nolimits_{p2g}(\my_i) \|.
          \end{align*}
    \item Normal consistency.
          \begin{align*}
              \texttt{NC} = & \frac{1}{2N_g}\sum_{i=1}^{N_g}  \left| \mn(\mx_i) \cdot \mn(\proj\nolimits_{g2p}(\mx_i)) \right| + \\ &  \frac{1}{2N_p}\sum_{i=1}^{N_p} \left| \mn(\my_i) \cdot \mn(\proj\nolimits_{p2g}(\my_i)) \right|.
          \end{align*}
    \item IOU is the volumetric intersection of two meshes divided by the volume of their union. To compute this metric, 100k points are sampled in the bounding box and are determined whether they are in or outside two meshes.
    \item F-Score is the harmonic mean between Precision and Recall. Precision is the percentage of points on $M_p$ that lie within distance $\tau$ to $M_g$, Recall is the percentage of points on $M_g$ that lie within distance $\tau$ to $M_p$.
          \begin{equation*}
              \texttt{F-Score} = \frac{2*\texttt{precision}*\texttt{Recall}}{\texttt{precision}+\texttt{Recall}}.
          \end{equation*}
\end{itemize}
We also compute the light field descriptor(LFD) to evaluate the perceptional similarity of the results to the ground-truth by following the setup of \cite{Chen2019}.  LFD is computed in the following way:  each generated shape is rendered from various views and results in a set of projected images, then each projected image is encoded using Zernike moments and Fourier descriptors.

\section{Network architecture} \label{appendix:net}
The number of network parameters for our network IMLSNet(7,7,1) reported in the paper is 4.6 M. The detailed setup of IMLSNet(7,6,1) and IMLSNet(7,7,1) is illustrated in the first and second rows of \cref{tab:network:channel}.  We also did an ablation study by setting the channel number of depth-2 and depth-3 octree nodes to 128 and reduced the network parameter size while achieving comparable performances as shown in \cref{tab:network:size}.  The networks are denoted by IMLSNet(6,6,1)$^\star$ and IMLSNet(7,7,1)$^\star$.

\begin{table}[t]
    \centering
    \scalebox{0.8}{
        \begin{tabular}{l|cccccc|c}
            \toprule
            \textbf{Network}       & 2   & 3   & 4   & 5  & 6  & 7  & size \\
            \midrule
            IMLSNet(7,6,1)         & 256 & 256 & 128 & 64 & 32 & 16 & 4.6M \\
            IMLSNet(7,7,1)         & 256 & 256 & 128 & 64 & 32 & 16 & 4.6M \\
            IMLSNet(7,6,1)$^\star$ & 128 & 128 & 128 & 64 & 32 & 16 & 1.5M \\
            IMLSNet(7,7,1)$^\star$ & 128 & 128 & 128 & 64 & 32 & 16 & 1.6M \\
            \bottomrule
        \end{tabular}
    }
    \caption{Network parameters of IMLSNets. Feature channel dimensions on each octree depth (from 2 to 7) are listed. }
    \label{tab:network:channel}
\end{table}

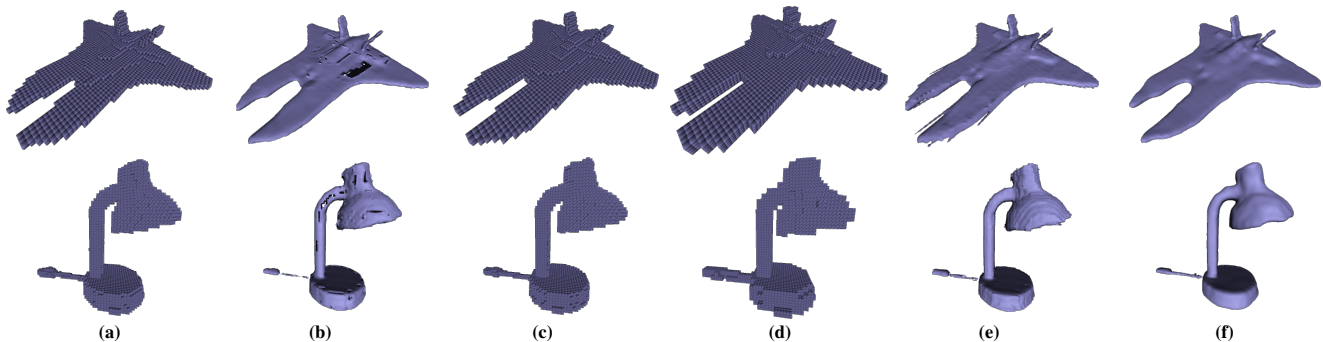
\begin{figure*}[t]
    \begin{overpic}[width=\linewidth]{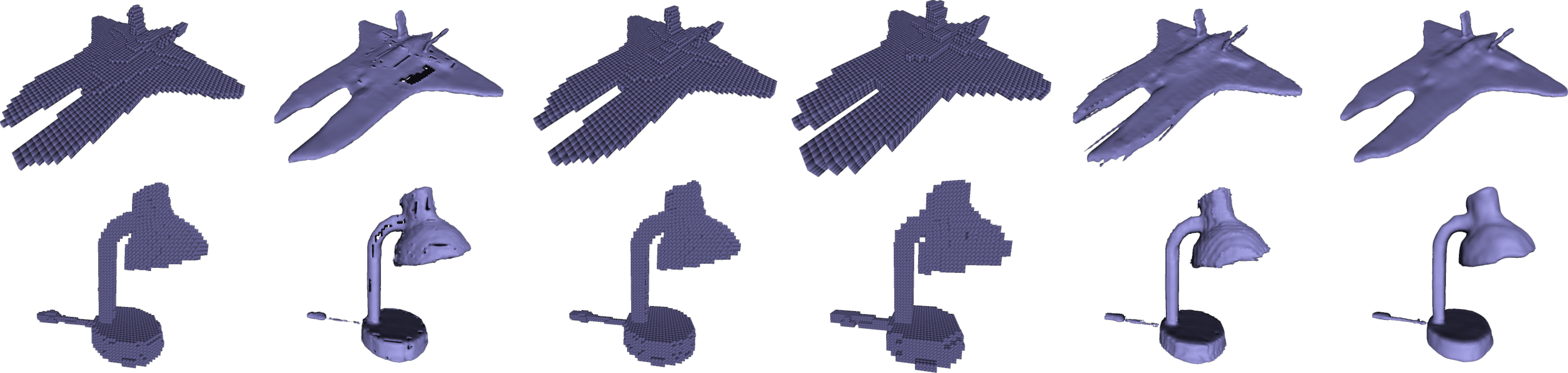}
        \put(7,-1.5){\scriptsize \textbf{(a)}}
        \put(23,-1.5){\scriptsize \textbf{(b)}}
        \put(40,-1.5){\scriptsize \textbf{(c)}}
        \put(58,-1.5){\scriptsize \textbf{(d)}}
        \put(74,-1.5){\scriptsize \textbf{(e)}}
        \put(92,-1.5){\scriptsize \textbf{(f)}}
    \end{overpic}
    \caption{(a): Non-empty finest octants predicated by the network based on the octree-aided deep local implicit function. (b): Reconstruction results from (a). (c): Ground-truth non-empty finest octants. (d): The expanded octree. (e): Reconstruction results based on (d). (f) Our IMLSNet results.} \label{fig:octfunc}
\end{figure*}

\begin{table}[t]
    \centering
    \scalebox{0.8}{
        \begin{tabular}{lcccc}
            \toprule
            \textbf{Network}       & \textbf{$\texttt{CD}_1$  $\downarrow$  } & \textbf{NC  $\uparrow$} & \textbf{IoU $\uparrow$} & \textbf{F-Score $\uparrow$} \\
            \midrule
            IMLSNet(7,6,1)         & 0.0310                                   & 0.9430                  & 0.9134                  & 0.9813                      \\
            IMLSNet(7,7,1)         & \textbf{0.0306}                          & \textbf{0.9440}         & \textbf{0.9135}         & \textbf{0.9833}             \\
            IMLSNet(7,6,1)$^\star$ & 0.0311                                   & 0.9425                  & 0.9129                  & 0.9814                      \\
            IMLSNet(7,7,1)$^\star$ & 0.0307                                   & 0.9434                  & 0.9132                  & 0.9827                      \\
            \bottomrule
        \end{tabular}
    }
    \caption{Quantitative evaluation of IMLSNet with different network settings on the task of 3D object reconstruction.}
    \label{tab:network:size}
\end{table}

\section{Octree-aided deep local implicit function} \label{appendix:discussion}
As discussed in Section 4, our ablation study shows that the octree-aided deep local implicit function has issues in obtaining complete zero-iso surfaces. We confirm this fact by examining the generated non-empty octants which cover the missing regions but the local implicit function does not generate the iso-surface in them.  \cref{fig:octfunc}-a illustrates these octants which cover the ground-truth non-empty octants (see \cref{fig:octfunc}-c) but fail to generate a complete shape (\cref{fig:octfunc}-b).   We speculate that the implicit function passes through the neighboring region which is not covered by the current finest octants. Based on this speculation, we split all the $d-1$-depth octants to expand the octree (see \cref{fig:octfunc}-d) and extract the surface via marching cubes. It turns out that the implicit surface appears in those regions (see \cref{fig:octfunc}-e), however, the reconstruction error is higher than our IMLSNet (\cref{fig:octfunc}-f).

\section{Evaluation of object reconstruction} \label{appendix:more}
\begin{figure*}[t]
    \begin{overpic}[width=\linewidth]{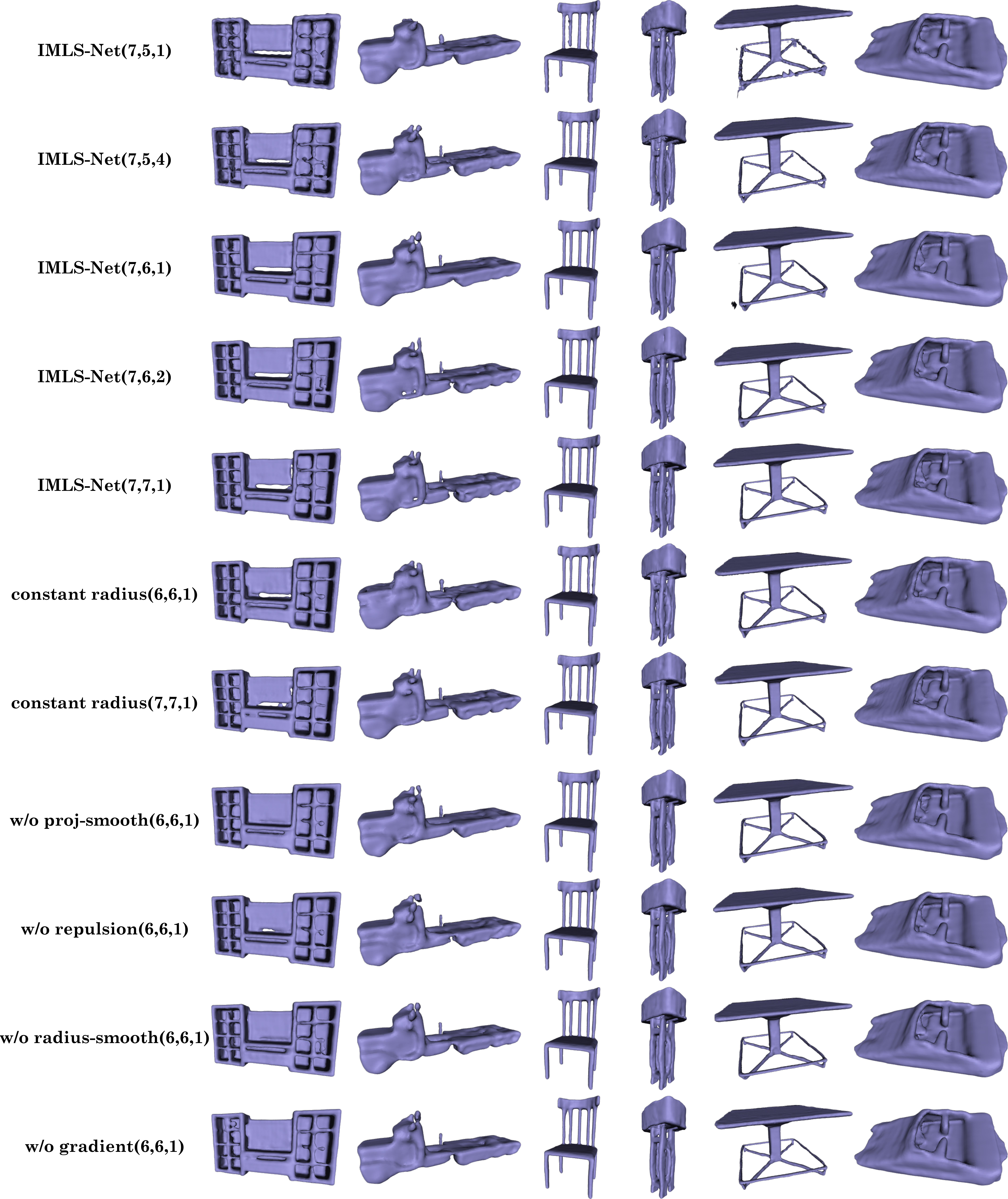}
    \end{overpic}
    \caption{More results of our ablation study on network settings. The inputs are the noisy point clouds (see \cref{fig:shape}).} \label{fig:ablation}
\end{figure*}

In \cref{tab:shape:cat}, we report the numerical metrics of the tasks of object reconstruction from point clouds for each shape category.  \cref{fig:more} presents more visual results reconstructed from our network.
All the evaluations demonstrate the superiority of our method over other approaches in terms of reconstruction accuracy and the capacity of recovering details and thin regions.  In \cref{fig:ablation} we present more results of our ablation study of different network settings.

\section{Robustness test on noise levels}\label{appendix:noise_ab}
We did a robustness test on the input noise. The network IMLSNet(7,7,1) and ConvOccNet were trained with noisy data whose Gaussian noise is with standard deviation $\delta = 5 \times 10^{-3}$. We add different noise levels ($\delta = 1 \times 10^{-3},3 \times 10^{-3},7.5 \times 10^{-3}$) to the test data of 13 shape classes and feed to our network and ConvOccNet for evaluating their performance. From \cref{tab:noise:ablation}, we can see with lower noise levels, our network always performs better than ConvOccNet. With a higher level noise ($\delta =7.5 \times 10^{-3}$), The network performance of both methods degrades gracefully, and our method still outperforms ConvOccNet.
\begin{table}[t]
    \centering
    \scalebox{0.75}{
        \begin{tabular}{cccccc}
            \toprule
            \textbf{Network} & \textbf{$\delta$}   & \textbf{$\texttt{CD}_1$  $\downarrow$  } & \textbf{NC  $\uparrow$} & \textbf{IoU $\uparrow$} & \textbf{F-Score $\uparrow$} \\
            \midrule
            IMLSNet          & $1.0\times 10^{-3}$ & \textbf{0.0288}                          & \textbf{0.9476}         & \textbf{0.9226}         & \textbf{0.9859}             \\
            ConvOccNet       & $1.0\times 10^{-3}$ & 0.0495                                   & 0.9349                  & 0.8573                  & 0.9442                      \\
            \midrule
            IMLSNet          & $3.0\times 10^{-3}$ & \textbf{0.0290}                          & \textbf{0.9473}         & \textbf{0.9219}         & \textbf{0.9857}             \\
            ConvOccNet       & $3.0\times 10^{-3}$ & 0.0439                                   & 0.9377                  & 0.8831                  & 0.9461                      \\
            \midrule
            IMLSNet          & $5.0\times 10^{-3}$ & \textbf{0.0306}                          & \textbf{0.9440}         & \textbf{0.9135}         & \textbf{0.9833}             \\
            ConvOccNet       & $5.0\times 10^{-3}$ & 0.0441                                   & 0.9383                  & 0.8842                  & 0.9421                      \\
            \midrule
            IMLSNet          & $7.5\times 10^{-3}$ & \textbf{0.0372}                          & \textbf{0.9291}         & \textbf{0.8754}         & \textbf{0.9705}             \\
            ConvOccNet       & $7.5\times 10^{-3}$ & 0.0536                                   & 0.9345                  & 0.8435                  & 0.9221                      \\
            \bottomrule
        \end{tabular}
    }
    \caption{Robustness test to noise. }
    \label{tab:noise:ablation}
\end{table}

\begin{table*}[t]
  \centering
  \scalebox{0.85}{
  \begin{tabular}{c|cccc|cccc}
  \toprule
  & \multicolumn{4}{c}{\textbf{$\texttt{CD}_1$  $\downarrow$}}     
  & \multicolumn{4}{c}{\textbf{NC  $\uparrow$}}    \\
  
  \textbf{Category} & O-CNN-C & IMLSNet points & ConvOccNet & IMLSNet &  O-CNN-C & IMLSNet points & ConvOccNet & IMLSNet  \\
  \midrule
  airplane        &0.0634	&0.0316	&0.0336  &\textbf{0.0245}		&0.9181	 &0.9292	 &0.9311	          &\textbf{0.9371} \\
  bench           &0.0646	&0.0356	&0.0352  &\textbf{0.0301}		&0.9136  &0.9194	 &0.9205	          &\textbf{0.9220} \\
  cabinet         &0.0709	&0.0375	&0.0461  &\textbf{0.0348}		&0.9411  &0.9505	 &\textbf{0.9561}	  &0.9546 \\
  car             &0.0765	&0.0419	&0.0750  &\textbf{0.0395}		&0.8668	 &0.8709	 &\textbf{0.8931}   &0.8820 \\	
  chair           &0.0664	&0.0383	&0.0459  &\textbf{0.0348}		&0.9407	 &0.9487	 &0.9427	          &\textbf{0.9503} \\
  display         &0.0655	&0.0339	&0.0368  &\textbf{0.0292}		&0.9598	 &0.9710	 &0.9677	          &\textbf{0.9732} \\
  lamp            &0.0667	&0.0367	&0.0595  &\textbf{0.0312}		&0.9111	 &0.9206	 &0.9003	          &\textbf{0.9218} \\
  speaker         &0.0729	&0.0413	&0.0632  &\textbf{0.0396}		&0.9363	 &0.9440	 &0.9387	          &\textbf{0.9473} \\
  rifle           &0.0617	&0.0300	&0.0280  &\textbf{0.0207}		&0.9320	 &0.9428	 &0.9293	          &\textbf{0.9433} \\
  sofa            &0.0657	&0.0350	&0.0414  &\textbf{0.0309}		&0.9492	 &0.9602	 &0.9579	          &\textbf{0.9631} \\
  table           &0.0663	&0.0360	&0.0385  &\textbf{0.0319}		&0.9461  &0.9599	 &0.9588	          &\textbf{0.9621} \\
  telephone       &0.0610	&0.0295	&0.0270  &\textbf{0.0229}		&0.9737  &0.9827   &0.9823	          &\textbf{0.9839} \\
  vessel          &0.0641	&0.0336	&0.0430  &\textbf{0.0271}		&0.9221	 &0.9280	 &0.9187	          &\textbf{0.9319} \\
  \midrule                                              
  mean            &0.0666 &0.0355 &0.0441  &\textbf{0.0306}   &0.9316  &0.9406    &0.9382           &\textbf{0.9440} \\
  \midrule
  bag    	        &0.0704 &0.0386 &0.0538 &\textbf{0.0351}  &0.9342	&0.9420	&0.9417	&\textbf{0.9455} \\
  bathtub	        &0.0663 &0.0378	&0.0526 &\textbf{0.0350}	&0.9478	&0.9599	&0.9537	&\textbf{0.9622} \\
  bed    	        &0.0720 &0.0428	&0.0608 &\textbf{0.0412}	&0.9192	&0.9246	&0.9119	&\textbf{0.9278} \\
  bottle 	        &0.0619 &0.0332	&0.0421 &\textbf{0.0279}	&0.9610	&0.9696	&0.9657	&\textbf{0.9708} \\
  pillow 	        &0.0631 &0.0340	&0.0548 &\textbf{0.0303}	&0.9652	&0.9743	&0.9660 &\textbf{0.9757} \\
  \midrule          
  mean            &0.0667 &0.0373 &0.0528 &\textbf{0.0339} &0.9455 &0.9541 &0.9478	&\textbf{0.9564} \\
  \bottomrule
  \end{tabular}
  }\\
\scalebox{0.7}{
\begin{tabular}{c|cccc|cccc}
	\toprule
	& \multicolumn{4}{c}{\textbf{IoU $\uparrow$}}     
	& \multicolumn{4}{c}{\textbf{F-Score $\uparrow$}}    \\
	
	\textbf{Category} & O-CNN-C & IMLSNet points & ConvOccNet & IMLSNet &  O-CNN-C & IMLSNet points & ConvOccNet & IMLSNet  \\
	\midrule	
	airplane 	&n/a	&n/a	&0.8485	 &\textbf{0.8910}	&0.8101	&0.9923	 &0.9653	   &\textbf{0.9918} \\
  bench    	&n/a	&n/a	&0.8298	 &\textbf{0.8480}	&0.7995	&0.9867  &0.9643	   &\textbf{0.9860} \\
  cabinet  	&n/a	&n/a	&0.9398	 &\textbf{0.9495}	&0.7887	&0.9833	 &0.9558	   &\textbf{0.9811} \\
  car      	&n/a	&n/a	&0.8858	 &\textbf{0.9052}	&0.7474	&0.9604	 &0.8490	   &\textbf{0.9521} \\
  chair    	&n/a	&n/a	&0.8709	 &\textbf{0.9033}	&0.7993	&0.9824	 &0.9387	   &\textbf{0.9815} \\
  display  	&n/a	&n/a	&0.9275	 &\textbf{0.9491}	&0.8109	&0.9935	 &0.9708	   &\textbf{0.9935} \\
  lamp     	&n/a	&n/a	&0.7840	 &\textbf{0.8583}	&0.7999	&0.9806	 &0.8910	   &\textbf{0.9785} \\
  speaker  	&n/a	&n/a	&0.9188  &\textbf{0.9450} &0.7789	&0.9676	 &0.8924	   &\textbf{0.9633} \\
  rifle    	&n/a	&n/a	&0.8459	 &\textbf{0.8856}	&0.8263	&0.9961	 &0.9799	   &\textbf{0.9962} \\
  sofa     	&n/a	&n/a	&0.9362	 &\textbf{0.9541}	&0.8052	&0.9886  &0.9531	   &\textbf{0.9873} \\
  table    	&n/a	&n/a	&0.8877	 &\textbf{0.9076}	&0.8013	&0.9875	 &0.9674	   &\textbf{0.9870} \\
  telephone	&n/a	&n/a	&0.9537	 &\textbf{0.9647}	&0.8252	&0.9978	 &0.9882	   &\textbf{0.9978} \\
  vessel   	&n/a	&n/a	&0.8663	 &\textbf{0.9140}	&0.8085	&0.9880	 &0.9313	   &\textbf{0.9868} \\
	\midrule                                          
	mean      &n/a  &n/a  &0.8842  &\textbf{0.9135}  &0.8001 &0.9850 &0.9421  &\textbf{0.9833} \\
	\midrule
  bag       &n/a  &n/a 	&0.9229	&\textbf{0.9461}	&0.7859	&0.9766	&0.9187	&\textbf{0.9745} \\  
  bathtub   &n/a  &n/a 	&0.8431	&\textbf{0.9079}	&0.8014	&0.9861	&0.9084	&\textbf{0.9851} \\
  bed       &n/a  &n/a 	&0.8612	&\textbf{0.9052}	&0.7727	&0.9661	&0.8985	&\textbf{0.9622} \\
  bottle    &n/a  &n/a 	&0.9468	&\textbf{0.9663}	&0.8282	&0.9916	&0.9515	&\textbf{0.9907} \\
  pillow    &n/a  &n/a 	&0.9354	&\textbf{0.9700}	&0.8219	&0.9948	&0.9005	&\textbf{0.9941} \\
	\midrule          
	mean      &n/a  &n/a  &0.9019 &\textbf{0.9391} &0.8020 &0.9831 &0.9155	&\textbf{0.9813} \\
	\bottomrule
\end{tabular}
}
  \caption{Quantitative evaluation of different networks on the test data of 13 shape classes and the full data of 5 unseen shape classes. }
  \label{tab:shape:cat}
  \end{table*}

\begin{figure*}[t]
    \begin{overpic}[width=\linewidth]{more}
        \put(2,-2){\scriptsize \textbf{Noisy input}}
        \put(13,-2){\scriptsize \textbf{ConvOccNet}}
        \put(27,-2){\scriptsize \textbf{IMLSNet}}
        \put(42,-2){\scriptsize \textbf{GT}}
        \put(53,-2){\scriptsize \textbf{Noisy input}}
        \put(65,-2){\scriptsize \textbf{ConvOccNet}}
        \put(78,-2){\scriptsize \textbf{IMLSNet}}
        \put(93,-2){\scriptsize \textbf{GT}}
    \end{overpic} \vspace{2mm}
    \caption{More results of object reconstruction from point clouds.} \label{fig:more}
\end{figure*}

\end{document}